\documentclass{article}

\usepackage{cite} 
\usepackage{graphics}
\usepackage{graphicx}
\usepackage[english]{babel}
\usepackage{amsmath}
\usepackage{amsthm}
\usepackage[table,xcdraw]{xcolor}
\usepackage[font=footnotesize]{caption}
\usepackage{epsfig}
\usepackage{xfrac}
\usepackage{nicefrac}
\usepackage{psfrag}
\usepackage{times}
\usepackage{balance} 
\usepackage{amssymb}
\usepackage{amsfonts}
\usepackage{bm}
\usepackage[bottom]{footmisc}
\usepackage{mwe}
\usepackage{color}
\usepackage{booktabs,multirow}
\usepackage{tabularx}
\usepackage{tikz}
\usetikzlibrary{patterns,arrows}
\usepackage{physics}
\usepackage{verbatim}
\usepackage{subcaption}
\usepackage[detect-all]{siunitx}
\usepackage{pifont}
\usepackage{enumitem}
\usepackage{hyperref}

\newcommand{\eg}{{\sl e.g., }}
\newcommand{\ie}{{\sl i.e., }}
\newcommand{\BfPara}[1]{{\noindent\bf#1.}\xspace}

\newcounter{descriptcount}
\newlist{enumdescript}{description}{1}
\setlist[enumdescript,1]{
before={\setcounter{descriptcount}{0}
\renewcommand*\thedescriptcount{\arabic{descriptcount}}},font={\bfseries\stepcounter{descriptcount}Q\thedescriptcount:}
}

\definecolor{color1}{RGB}{228,26,28}
\definecolor{color2}{RGB}{55,126,184}
\definecolor{color3}{RGB}{77,175,74}
\definecolor{color4}{RGB}{152,78,163}
\definecolor{color5}{RGB}{255,127,0}
\definecolor{color6}{RGB}{200,200,200}

\newcommand{\cmark}{\ding{51}}
\newcommand{\xmark}{\ding{55}}

\usepackage{hyperref}

\usepackage[accepted]{icml2025}

\icmltitlerunning{On‑Device LLM for Context‑Aware Wi‑Fi Roaming}

\begin{document}

\twocolumn[
\icmltitle{On‑Device LLM for Context‑Aware Wi‑Fi Roaming}




\begin{icmlauthorlist}
\icmlauthor{Ju-Hyung Lee}{nokia}
\icmlauthor{Yanqing Lu}{nokia,usc}
\icmlauthor{Klaus Doppler}{nokia}
\end{icmlauthorlist}

\icmlaffiliation{nokia}{Nokia Technologies, Sunnyvale, CA, USA}
\icmlaffiliation{usc}{Department of Computer Science, University of Southern California, Los Angeles, CA, USA}

\icmlcorrespondingauthor{Ju-Hyung Lee}{juhyung.lee@nokia.com}

\icmlkeywords{Machine Learning, ICML, On-Device LLM, LLM, Wi-Fi, Roaming, Handover, Cross-Layer Optimization, PHY, MAC, Language Model, Fine Tuning, Post Training, Context‑Aware Wi‑Fi Roaming, Cross‑Layer Control, Edge AI, In‑Context Learning, Chain‑of‑Thought Prompting, Parameter‑Efficient Fine‑Tuning, LoRA, Quantization, Preference Optimization, PHY/MAC, Wireless Communications
}

\vskip 0.3in
]



\printAffiliationsAndNotice{}  

\begin{abstract}
    Roaming in Wireless LAN (Wi-Fi) is a critical yet challenging task for maintaining seamless connectivity in dynamic mobile environments. Conventional threshold‑based or heuristic schemes often fail, leading to either \emph{sticky} or \emph{excessive} handovers. We introduce the first \emph{cross‑layer} use of an on‑device large language model (LLM): high‑level reasoning in the application layer that issues real‑time actions executed in the PHY/MAC stack. The LLM addresses two tasks: (\emph{i}) \textit{context‑aware AP selection}, where structured prompts fuse environmental cues (e.g., location, time) to choose the best BSSID; and (\emph{ii}) \textit{dynamic threshold adjustment}, where the model adaptively decides \emph{when} to roam. To satisfy the tight latency and resource budgets of edge hardware, we apply a suite of optimizations—chain‑of‑thought prompting, parameter‑efficient fine‑tuning, and quantization. Experiments on indoor and outdoor datasets show that our approach surpasses legacy heuristics and DRL baselines, achieving a strong balance between roaming stability and signal quality. These findings underscore the promise of application‑layer LLM reasoning for lower‑layer wireless control in future edge systems.

\end{abstract}
\section{Introduction} \label{sec:intro}

Wi-Fi roaming is a critical operation for maintaining seamless wireless connectivity as users move through physical environments. Traditionally, roaming logic has assumed relatively stable topologies—stationary access points (APs) and predictable client mobility. However, emerging scenarios are rapidly redefining this assumption. Automotive Wi‑Fi and device‑to‑device (D2D) “mobile‑AP” scenarios all feature mobility on \emph{both} the client and AP sides, creating highly dynamic topologies that demand instantaneous, context‑aware decisions at the wireless edge.

Despite decades of refinement, current roaming mechanisms still rely heavily on threshold-based triggers—such as scanning when RSSI drops below $-70$~dBm—and static handover logic that fails to adapt to varying conditions. Such strategies often result in either \textit{sticky handovers} (where devices cling to weak links) or \textit{excessive handovers} (leading to instability and overhead). These issues are exacerbated in modern use cases, where wireless conditions change rapidly across space and time.

\begin{figure}[!t]
\centering
\includegraphics[width=1.\linewidth]{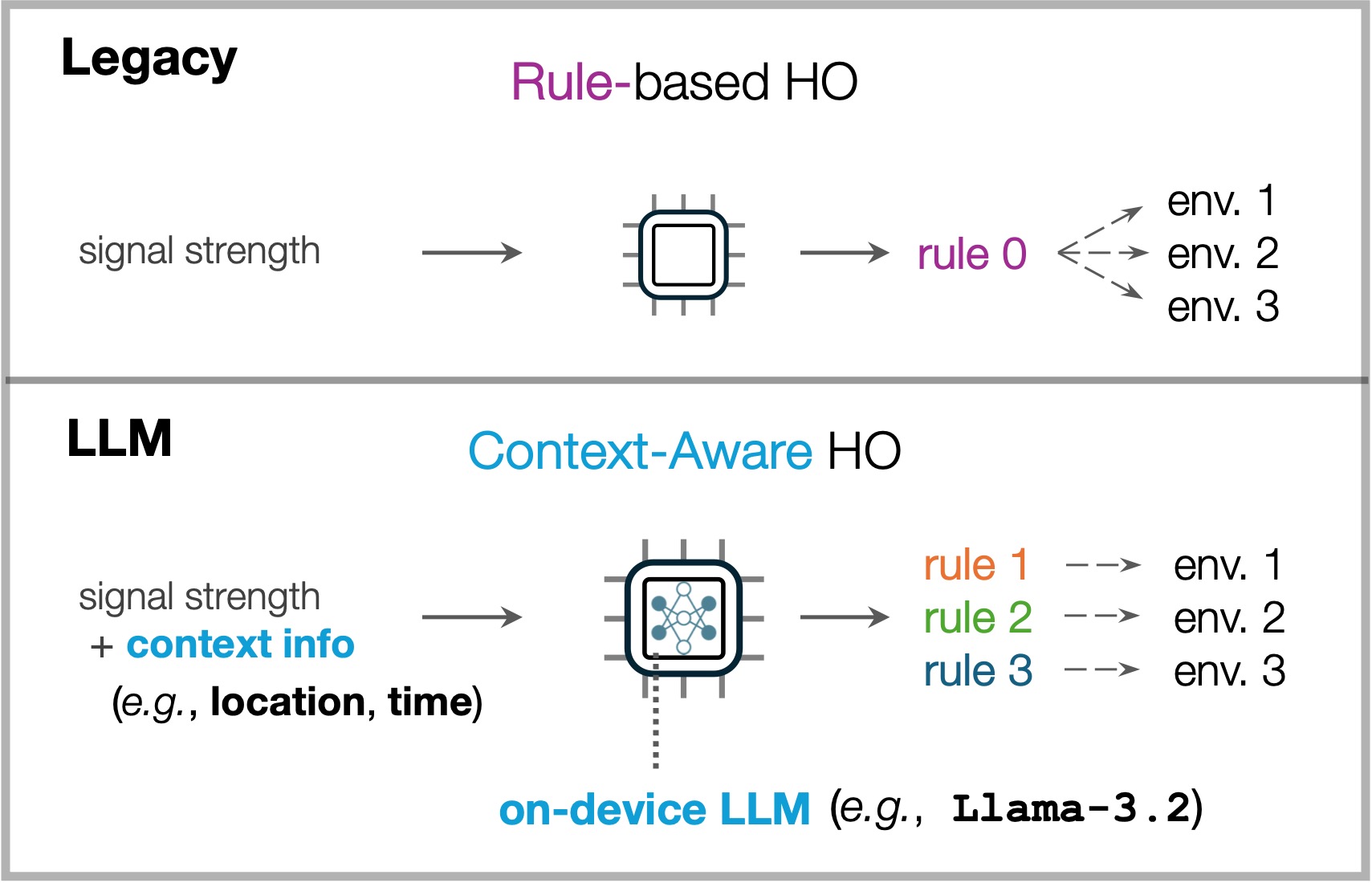}
\caption{Cross-layer control via on-device LLM: Rule‑based handover (legacy) vs. on‑device LLM context‑aware handover.}
\label{fig:overview_CHO}
\vspace{-1.em}
\end{figure}

Making intelligent roaming decisions under these constraints is inherently difficult. The system must reason over noisy, high-dimensional signals such as RSSI patterns, user location, time of day, and device state—all under tight latency budgets (typically within 10--100~ms). Prior approaches based on supervised learning or deep reinforcement learning (DRL) have shown promise but suffer from limited generalization, requiring task-specific retraining and extensive engineering to adapt to new environments.

In this work, we propose a new paradigm: using large language model (LLM) as adaptive roaming agents deployed directly at the wireless edge. While LLMs have recently been explored for cloud‑level network management~\cite{wu2024netllm}, \textbf{\emph{this work is the first to deploy an LLM \emph{on-device} for real-time control at the lower layers of a wireless system, specifically at the PHY/MAC level}}. By leveraging the LLM’s in-context reasoning capabilities, we enable it to interpret structured prompts that encode real-time situational context—such as RSSI, location, and time—and make intelligent roaming decisions locally, {without external coordination or retraining}.

\noindent\textbf{Problem scope.}  
We study two concrete tasks:
\begin{enumerate}[label=\textbf{T\arabic*}), leftmargin=*]
  \item \textit{Context‑aware AP selection}~(\S\ref{sec:task1}): choose the optimal BSSID given current context;
  \item \textit{Dynamic threshold adjustment}~(\S\ref{sec:task2}): adaptively decide \emph{when} to trigger roaming.
\end{enumerate}

\noindent\textbf{Contributions.}
\begin{itemize}
  \item \textbf{Cross‑layer Wireless Control via On-Device LLM.}  We demonstrate the first on‑device LLM that reasons in the application layer while issuing real‑time actions in the PHY/MAC stack.
  \item \textbf{Edge‑efficient LLM pipeline.}  Through prompt design, post-training, and quantization, we trim memory and compute cost, pushing inference toward real‑time with only marginal accuracy loss.
  \item \textbf{Comprehensive evaluation.}  Indoor and outdoor experiments show our approach outperforms legacy and DRL baselines, striking a strong trade‑off between roaming stability and link quality
  \item \textbf{Practical insights \& Real-world demo.}  We distill practical guidelines for deploying LLMs as edge‑level wireless controllers (\S\ref{sec:discussion}) and validate our approach through practical demonstrations (\S\ref{sec:demo})\footnote{A full demonstration video and inference code are available at \href{https://github.com/abman23/on-device-llm-wifi-roaming}{github.com/abman23/on-device-llm-wifi-roaming}.}.  
\end{itemize}

We begin in Section~\ref{sec:background} by reviewing the fundamentals of Wi-Fi roaming and describing how LLMs can act as real-time, context-aware decision-making agents in wireless systems.


\section{Background} \label{sec:background}

\subsection{Wi-Fi Roaming and Mobility Management}

In Wi-Fi networks, \textit{roaming} refers to the process by which a client device (\eg a smartphone or laptop) switches its connection from one AP to another as the user moves. These APs typically belong to the same extended service set (ESS), providing overlapping coverage areas. For instance, as a user walks through a building, the device must determine when it should disconnect from the current AP and connect to another AP to maintain a strong wireless connection and avoid dropped connections or degraded signal quality.

Roaming is a Layer~2 operation governed by the \texttt{IEEE 802.11} standard, playing a crucial role in ensuring seamless connectivity across dynamic environments such as office buildings, university campuses, and public transit stations.

Most client devices continuously monitor the received signal strength indicator (RSSI) of their currently associated AP. When the RSSI falls below a predefined threshold, known as \texttt{scanRSSI} (typically set around $-70$~dBm), the device initiates a roaming procedure. After crossing this threshold, the device actively scans for candidate APs, evaluating them based on factors such as RSSI, channel congestion, and PHY-layer capabilities. It then selects a new AP if this AP offers significantly better link quality, commonly defined by a relative RSSI improvement threshold (\eg at least $+8$~dB during active data transmission or $+12$~dB during idle periods).

\begin{figure}[!h]
    \centering  
    \begin{subfigure}[t]{.48\linewidth}
        \centering  
        \includegraphics[width=\linewidth]{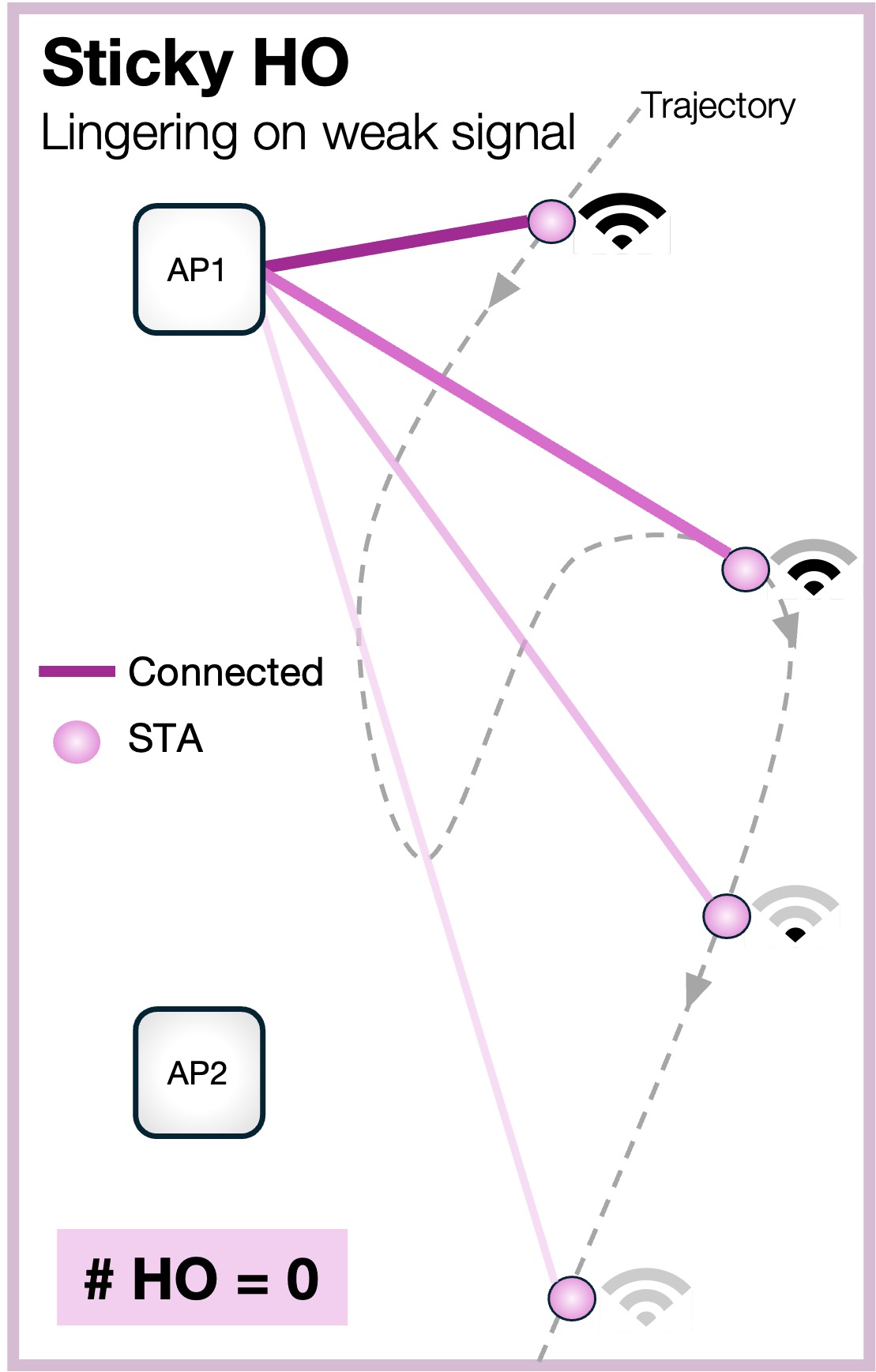}
        \caption{Sticky handover.}
    \end{subfigure}\hspace*{.015\textwidth}%
    \begin{subfigure}[t]{.48\linewidth}
        \centering  
        \includegraphics[width=\linewidth]{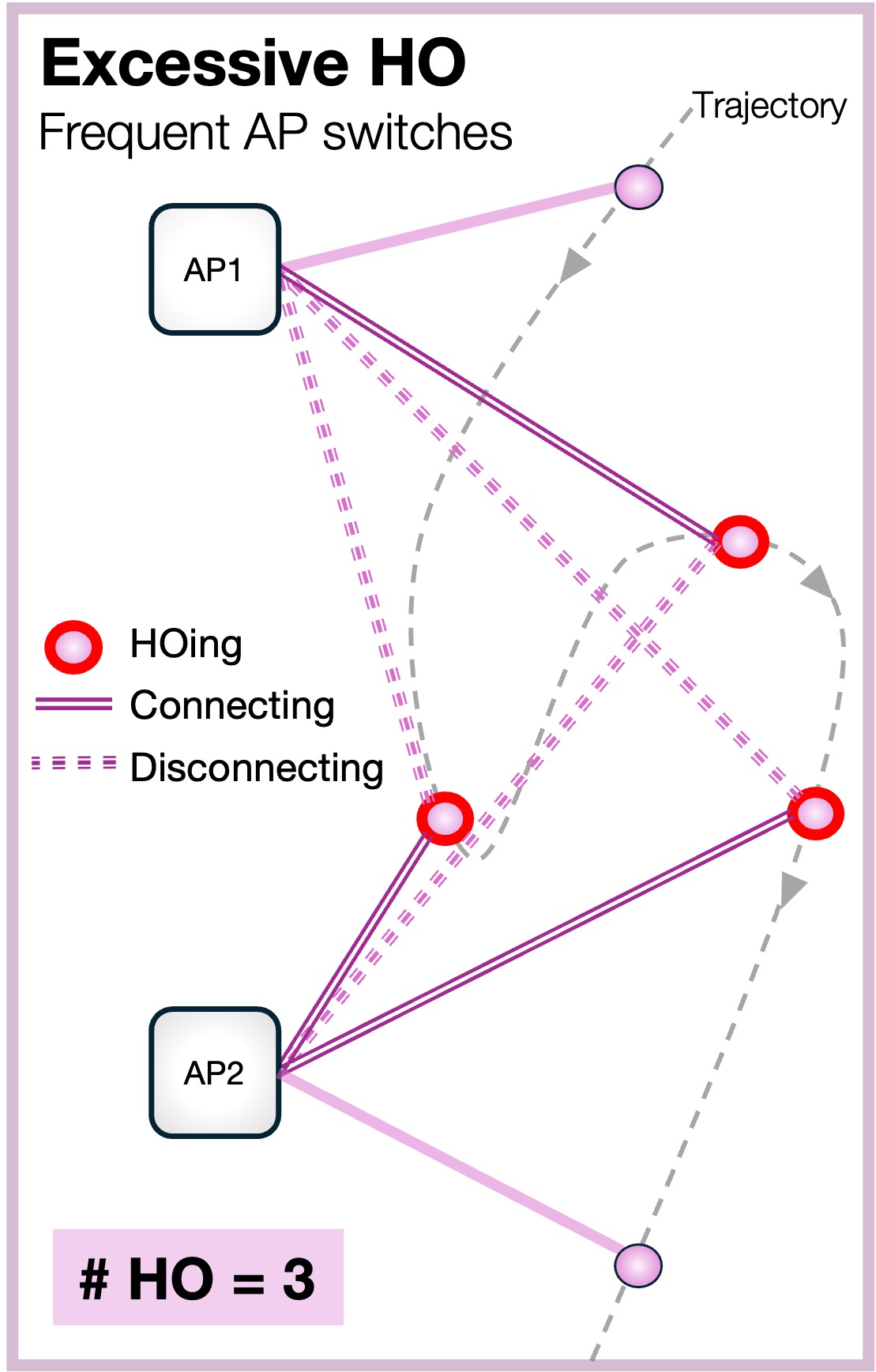}
        \caption{Excessive handover.}
    \end{subfigure}
\centering
\caption{Common failure modes of rule-based Wi-Fi roaming.}
\label{fig:challenge_handover}
\end{figure}

\begin{table*}[!h]   
\centering
\resizebox{1.4\columnwidth}{!}{
\centering
\begin{tabular}{l l l}
\toprule
\textbf{Aspect} & \textbf{Conventional ML / DRL} & \textbf{LLMs as Agents} \\
\midrule
Input Format & Fixed feature vectors & Flexible, multimodal prompts \\
\cmidrule(lr){1-3}
Adaptability & Retraining required for new settings & Zero-/few-shot generalization \\
\cmidrule(lr){1-3}
Reasoning Style & Implicit via model weights & Explicit, interpretable (\eg CoT) \\
\cmidrule(lr){1-3}
Context Integration & Hand-crafted, limited scope & Natural support for rich context \\
\cmidrule(lr){1-3}
Deployment & Efficient but narrow & Powerful, yet compute-heavy \\
\bottomrule
\end{tabular}
}
\caption{Comparison: Conventional AI vs. LLM as Decision-Making Agents}
\label{tab:llm_vs_ml}
\end{table*} 



Despite their simplicity, fixed RSSI thresholds and rule-based roaming logic frequently lead to suboptimal outcomes. As shown in Fig.~\ref{fig:challenge_handover}, poorly tuned thresholds and static decision rules often produce two problematic scenarios:
\begin{itemize}
    \item \textbf{Sticky handovers:} The device remains connected to a weakening AP, causing deteriorating throughput, increased latency, and poor user experience.
    \item \textbf{Excessive handovers:} The device frequently initiates scanning and switches between APs, causing disruptions in the connection, increased signaling overhead, lowered throughput, and degraded user experience.
\end{itemize}

Effectively managing roaming thus requires balancing two conflicting objectives: maintaining high signal quality while minimizing handover frequency. To quantitatively evaluate roaming strategies, we consider two key metrics:
\begin{itemize}
    \item \textbf{AvgRSSI}: The average RSSI experienced by the client (Station, STA) throughout the test period, reflecting the overall connection quality (higher is better).
    \item \textbf{\#HO}: The total number of handovers triggered during the evaluation period, indicating roaming stability or the overhead associated with frequent AP switching (lower is better).
\end{itemize}

A well-designed roaming policy must intelligently and dynamically balance these two metrics based on real-time environmental context and user requirements.

\subsection{LLMs as Decision-Making Agents}

LLMs have shown impressive generalization across diverse tasks in language, vision, reasoning, and planning. More recently, they have been explored as \textit{decision-making agents} capable of operating in structured environments such as robotics and human-computer interaction \cite{irpan2022do, huang2022languagemodelszeroshotplanners}—domains that demand not only prediction but also adaptive control.

In wireless systems, especially at the lower protocol layers (\eg PHY/MAC), such adaptive capabilities remain underexplored. Traditional methods in tasks like roaming, scheduling, and interference management typically rely on fixed heuristics or conventional machine learning models. Reinforcement learning approaches (\eg PPO) have been applied in wireless contexts; however, these methods often demand specialized model designs, dense reward signals, and substantial retraining efforts when deployed in new environments or tasks \cite{lacava2024programmable, lee2024handover, wilhelmi2024machinelearningwifi}.

LLMs offer a fundamentally different paradigm. They enable \textit{in-context learning}, allowing the model to make structured decisions based on prompt-based inputs without task-specific retraining. This adaptability makes LLMs well-suited for dynamic wireless settings, where behavior must generalize across diverse users, locations, and temporal conditions. LLMs can naturally process heterogeneous input modalities—such as signal strength, location, time of day, and battery state—using structured prompts, and can output decisions accompanied by interpretable reasoning traces (\eg via chain-of-thought prompting).

These capabilities position LLMs not simply as prediction or classification models but as \textit{context-aware agents} capable of integrating high-dimensional contextual information for adaptive wireless control. Although LLM inference occurs in the application layer, their outputs directly influence PHY/MAC layer parameters—such as selecting a target BSSID or adjusting roaming thresholds. This represents an emerging trend of deploying \emph{application-layer AI for lower-layer wireless control}. Table~\ref{tab:llm_vs_ml} highlights key differences between conventional AI methods and LLM-based approaches in this scenario.

Despite their promise, deploying LLMs for real-time wireless control introduces practical challenges, particularly on edge devices. Strict latency constraints (typically within 100~ms) combined with limited computational and memory resources pose significant hurdles \cite{xu2024ondevicelanguagemodelscomprehensive}. In this work, we overcome these constraints through a combination of model-level optimizations (quantization and parameter-efficient fine-tuning) and task-level adaptations designed to minimize inference overhead. These strategies enable the deployment of adaptive, context-aware LLMs capable of efficient, real-time operation directly on-device.

\section{Task (1): Context-Aware AP Choice by LLM} \label{sec:task1}
Our first objective is to determine whether a LLM can improve Wi-Fi roaming by selecting the optimal \emph{basic service set identifier} (BSSID) using real-time context (\eg device location and time).

\subsection{Problem Statement: Best BSSID Selection}
When multiple APs are available, the device must select BSSID to roam to. The “best” node is typically the one offering the most favorable channel connection (\eg, strong RSSI). In dynamic environments, the optimal choice can depend not only on RSSI, but also other context information. For example, site-specific information might hint at which AP has better coverage, while time of day and user mobility might tells the network congestion patterns. Our task is to predict, at any given moment, which AP the client should associate with to maximize handover performance.

\BfPara{Challenging Scenarios} 
As shown in Figure~\ref{fig:challenge_handover}, traditional roaming mechanisms often exhibit two critical shortcomings: \emph{sticky handovers}, where the device remains connected to a suboptimal AP despite significant signal degradation (left panel, zero handovers), and \emph{frequent handovers}, where the device constantly re-associates with different APs (center panel, multiple handovers), leading to instability and increased overhead. These scenarios degrade user experience via dropped connections, lower throughput, and higher latency.

\subsection{Approach: LLM-based Decision with Contextual Information}
We employ a LLM to evaluate available APs based on rich situational data. Specifically, the model ingests inputs such as the current and neighboring AP RSSI, location, time of day, and historical throughput. These contextual features may be encoded  in a structured prompt or feature vector that the LLM can interpret.  Unlike conventional AI solutions, which often require significant re-training or custom engineering  to handle new context signals, an LLM can flexibly process additional inputs by extending the prompt format. The LLM then predicts which BSSID the STA should join next (or remain connected to, if already optimal). 
By going beyond a simple “highest RSSI wins” legacy, the model can learn nuanced rules  (for instance, staying on a slightly weaker AP if it provides better backhaul,  or deferring handover if a signal dip is only transient), thereby adapting more effectively to diverse and evolving environments.

\BfPara{Prompting Engineering} We adopt \emph{chain‐of‐thought} (CoT) \cite{wei2022chain} prompting and examine how few‐shot examples affect the LLM’s decision‐making. Specifically, we provide the model with situational context (current RSSI, neighboring AP RSSIs, user location, time of day, battery state) and a small number of labeled examples—ranging from zero to five “shots.” In the CoT variant, each example includes a concise reasoning trace before the final decision, whereas the non‐CoT variant supplies only the final label or action.

While CoT improves zero-/few-shot reasoning, we still need domain adaptation via supervised fine-tuning. Then, to align the model’s decisions with user preferences, we apply \emph{direct preference optimization} (DPO) \cite{rafailov2023dpo}.

\BfPara{Fine-Tuning} We fine-tune the pre-trained LLM on real Wi-Fi roaming log data. We adopt LoRA \cite{hu2022lora} for \emph{parameter-efficient fine-tuning} (PEFT). Using LoRA, we inject small adaptation parameters and even fine-tune on a quantized model to save memory. This adaptation trains the LLM to understand Wi-Fi-specific cues (e.g., what RSSI patterns precede a disconnection) and to output the best AP choice accordingly. The lightweight nature of LoRA fine-tuning allows us to iterate and improve the model without needing enormous compute resources.

\BfPara{Preference-Based Learning} We further refine the model through preference optimization. For instance, logs or user feedback indicating which handover decisions yield better outcomes (e.g., higher throughput, fewer drops) are treated as preferred. The LLM is then optimized to prioritize these decisions via DPO, which directly tunes the model toward favored behaviors. During this process, the LLM explores candidate AP choices, scores them with a learned function, and iteratively updates its policy to favor those leading to better results. The end result is an LLM that not only produces valid AP selections but also aligns with real-world performance objectives.

\subsection{Evaluation} \label{sec:evaluation}
We evaluate the Task (1): Context-Aware AP Choice by comparing our LLM-based approach against several baseline roaming strategies in a variety of Wi-Fi roaming scenarios.
To evaluate roaming behavior, we use two key metrics:
\begin{align}
\text{\textit{AvgRSSI}} &= \frac{1}{T} \sum_{t=1}^{T} \text{RSSI}_t, \\    
\text{\textit{\# HO}} &= \sum_{t=2}^{T} \mathbb{I}[\text{BSSID}_t \ne \text{BSSID}_{t-1}],
\end{align}
\noindent
Here, $T$ denotes the number of time steps, $\text{RSSI}_t$ is the received signal strength at time $t$, and $\text{BSSID}_t$ is the AP the device is associated with. \textbf{\textit{AvgRSSI}} reflects connection quality, while \textbf{\textit{\# HO}} captures roaming stability. These two objectives often conflict: reducing handovers may hurt signal quality, and maximizing RSSI may cause unnecessary roaming. A well-designed roaming policy must intelligently balance these competing goals based on context.

The baselines for comparison are as follows:
\begin{itemize}
    \item {\textsf{Heuristic}}: randomly selects an AP when RSSI falls below \texttt{scanRSSI}, triggering a roam.
    \item {\textsf{Legacy}}: chooses the AP with the highest RSSI when RSSI is below \texttt{scanRSSI}, thereby triggering a roam.
    \item {\textsf{opt-HO} \& \textsf{opt-RSSI}}: {opt-HO} selects the AP that minimizes {\textit{\# HO}}, while {opt-RSSI} chooses the AP that maximizes \textit{{AvgRSSI}}, over the test sequence. Both methods achieve global optimality via exhaustive search.
    \item {\textsf{PPO}} \cite{schulman2017ppo}: is a DRL agent trained for AP selection that takes Wi-Fi measurements and contextual information as input, then outputs the target AP for roaming.
    \item {\textsf{LLM}}: is our proposed LLM-based method, which leverages context-aware prompting to determine the optimal AP.
\end{itemize}

Detailed experimental setups and parameter settings are provided in Table~\ref{tab:setup} in Appendix~\ref{sec:setup}.

\subsection{Experiments}

\begin{figure}[!h]
    \centering
    \includegraphics[width=1.0\columnwidth]{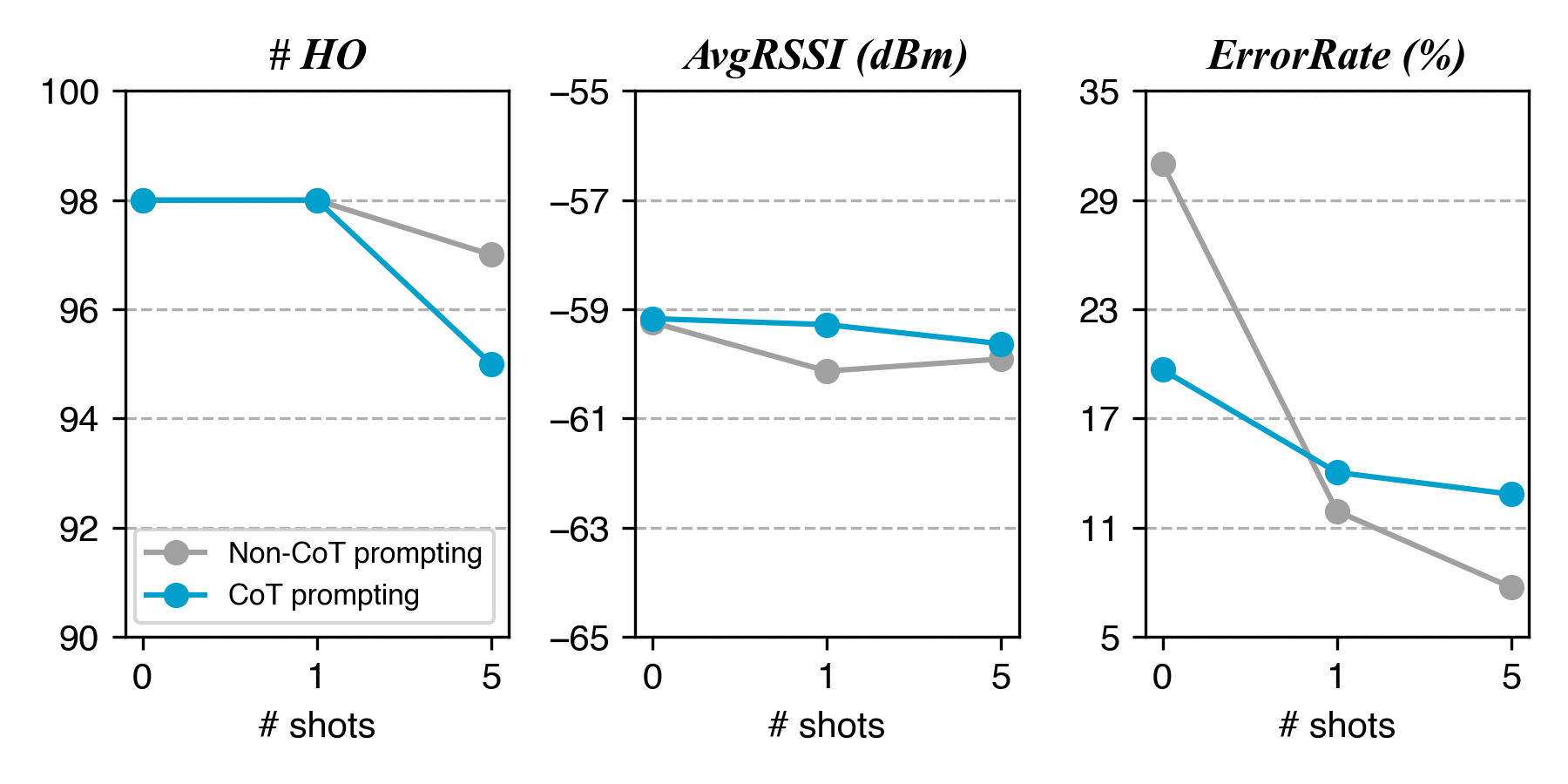}
    \caption{Impact of prompt engineering.}
    \label{fig:prompt_engineering}
\end{figure}

\BfPara{Effect of Prompt Engineering}
Figure~\ref{fig:prompt_engineering} compares CoT prompting with a simpler, non-CoT approach across 0-, 1-, and 5-shot scenarios, highlighting three main findings: \emph{i)}~CoT consistently reduces handover count, indicating fewer unnecessary switches; \emph{ii)}~CoT maintains a stronger average signal (by about 1--2\,dB); and \emph{iii)}~CoT begins with a lower error rate at 0-shot, although non-CoT catches or surpasses it at 5-shot. Overall, CoT proves especially beneficial in few-shot contexts, whereas non-CoT capitalizes more effectively on additional examples.

\BfPara{Effect of Post-Training}
To reduce computational overhead without compromising accuracy, we adopt PEFT via LoRA. As shown in Table~\ref{table:ablation_PEFT}, with a suitable configuration (\eg a learning rate of $2\times10^{-4}$, batch size of $2$, and LoRA rank of $128$), the model achieves about $85 \%$ accuracy to test set for opt-HO-global optimal (exahasutive) results for \# of roaming, while using only about $20\%$ of VRAM and $30\%$ of the GPU-hours required for full training. This efficiency–accuracy balance makes LoRA an attractive choice for computational heavy LLM fine-tuning, so that is used for our post-training.

\begin{table}[!h]   
\centering
\resizebox{.95\columnwidth}{!}{
\centering
  \begin{tabular} {c c c c | c}
\toprule[1pt]
\textbf{Learning Rate} & \textbf{Batch Size} & \textbf{Rank} (LoRA) & \textbf{Quantization} & \textbf{\textit{Accuracy}}~(\%)\\
\midrule
1e-5 & 1 & 32 & \xmark & {57.89}\\
2e-4 & 1 & 32 & \cmark & 68.42\\  
2e-4 & 1 & 32 & \xmark & 69.47\\
2e-4 & 2 & 32 & \xmark & {78.95}\\
2e-4 & 1 & 128 & \xmark & 78.95\\
2e-4 & 2 & 128 & \xmark & {85.26}\\
\bottomrule[1pt]
  \end{tabular}

}
\caption{Performance across various PEFT configurations. Accuracy is the percentage of LLM-selected APs matching test labels generated by opt-HO, which minimizes handovers over the time sequence (Base model: \texttt{Llama3.1}-8B \cite{grattafiori2024llama}).} 
\label{table:ablation_PEFT}
\vspace{-0.5em}
\end{table} 

Table~\ref{table:posttrain_task1} compares different post-training methods. 
While supervised fine-tuning (SFT) improves the average RSSI, it does not reduce handovers or the error rate. By contrast, combining SFT with DPO lowers the error rate significantly (to $12.66\%$), suggesting a more balanced outcome. 
Odds ratio preference optimization (ORPO)~\cite{hong2024orpo} further cuts handovers but drives up the error rate to $33\%$, underscoring a trade-off. 
Nevertheless, in scenarios prioritizing fewer roam events over higher error tolerance (\eg when a fallback mechanism is available), ORPO may still be advantageous.

\begin{table}[!h]   
\centering
\resizebox{.9\columnwidth}{!}{
\centering
\begin{tabular} {c|cc|c}
\toprule[1pt]
\textbf{Method} & \textbf{\textit{\# HO}} & \textbf{\textit{AvgRSSI}}~(dBm) & \textbf{\textit{ErrorRate}}~(\%)\\
\midrule
No FT & 35.5 & -56.48 & 22.83\\
SFT & 35.5 & -55.91 & 22.83\\
SFT+DPO & 34.5 & -56.53 & 12.66\\
\textbf{ORPO} & 33.0 & -55.91 & 33.00\\
\bottomrule[1pt]
\end{tabular}
}
\caption{Impact of post-training methods. \emph{ErrorRate} denotes the fraction of invalid AP selections (\ie when the LLM chooses an AP that is unavailable or has an RSSI below \texttt{scanRSSI}). The base model used is \texttt{Llama3.1}-8B \cite{grattafiori2024llama}.}
\label{table:posttrain_task1}
\end{table} 

\begin{figure*}[!h]
\centering
\includegraphics[width=.5\linewidth]{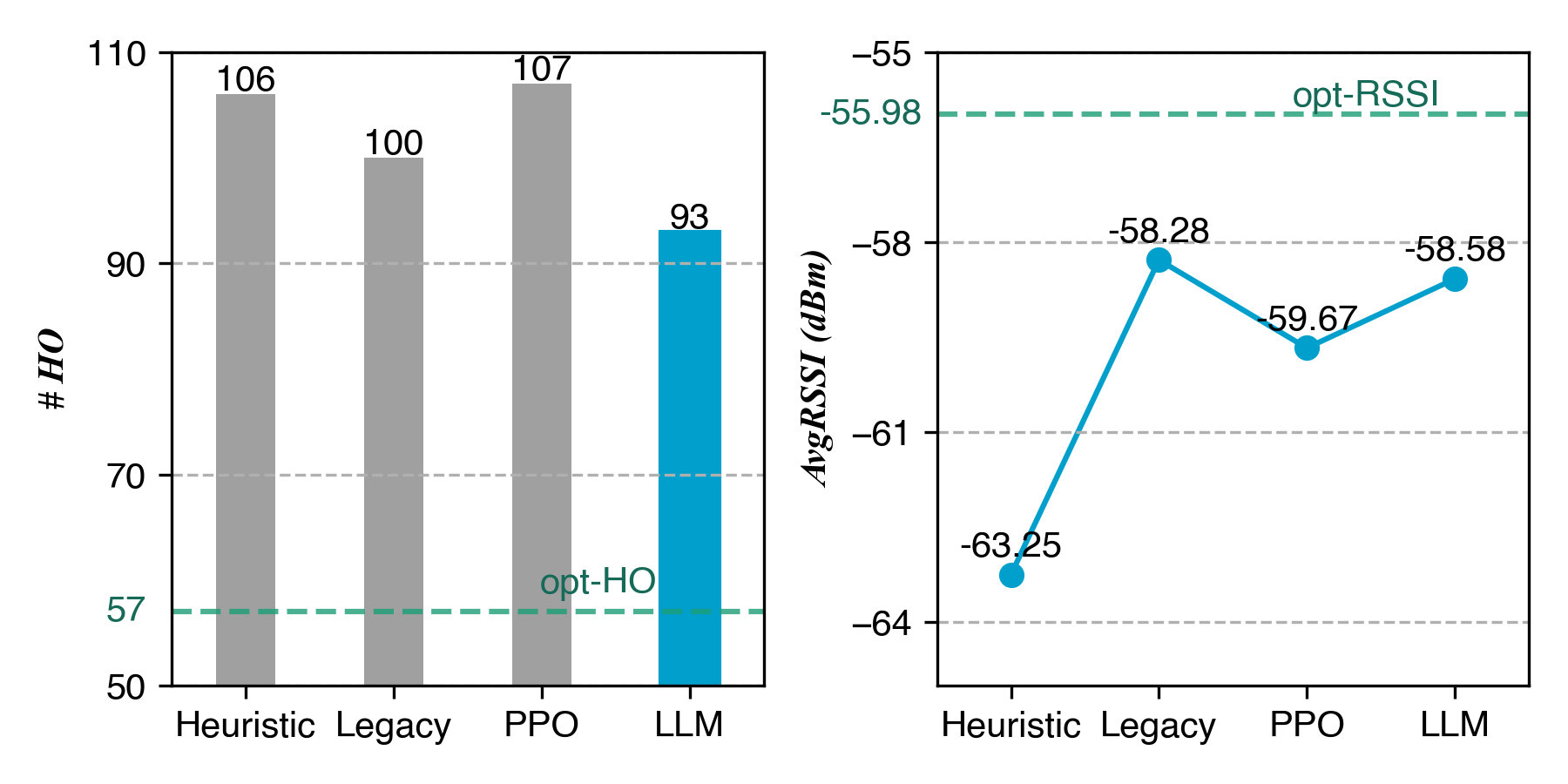}
    \caption{Comparison of \textbf{\textit{\# HO}} (left) and \textbf{\textit{AvgRSSI}} (right) for each method for best BSSID selection (Task~1).}
\label{fig:comparison_task1}
\vspace{-1.em}
\end{figure*}

\BfPara{Comparison: Legacy vs. DRL vs. LLM}
We next evaluate our LLM approach against textsf{Legacy}, 
an offline DRL method based on proximal policy optimization (PPO), and two global optimal results that minimize handovers (\textsf{opt-HO}) or maximize signal strength (\textsf{opt-RSSI}). Figure~\ref{fig:comparison_task1} illustrates the trade-offs between handover counts (left) and received signal strength (right) for each strategy. 
The \textsf{LLM} (93 handovers) outperforms \textsf{Heuristic} (106), \textsf{Legacy} (100), and \textsf{PPO} (107) in reducing unnecessary roaming, though it does not reach the minimal handover level of \textsf{opt-HO} (57). 
Meanwhile, it preserves a higher average signal ($-58.58\,\mathrm{dBm}$) than \textsf{Heuristic} ($-63.25$), \textsf{PPO} ($-59.67$), or \textsf{opt-HO} ($-63.81$), and comes reasonably close to the \textsf{Legacy} method ($-58.28$) and \textsf{opt-RSSI} ($-55.98$).

Overall, these results demonstrate that each extreme heuristic sacrifices one metric to excel at the other: opt-HO drastically lowers handovers but offers weak signal quality, whereas opt-RSSI achieves excellent signal strength at the cost of excessive handovers. By contrast, the LLM balances both goals, achieving fewer handovers than conventional baselines and maintaining a robust connection, highlighting the benefits of a context-aware decision-making framework.

\subsection{Limitations} \label{sec:limitation}
\begin{figure}[!h]
\centering
\includegraphics[width=0.8\linewidth]{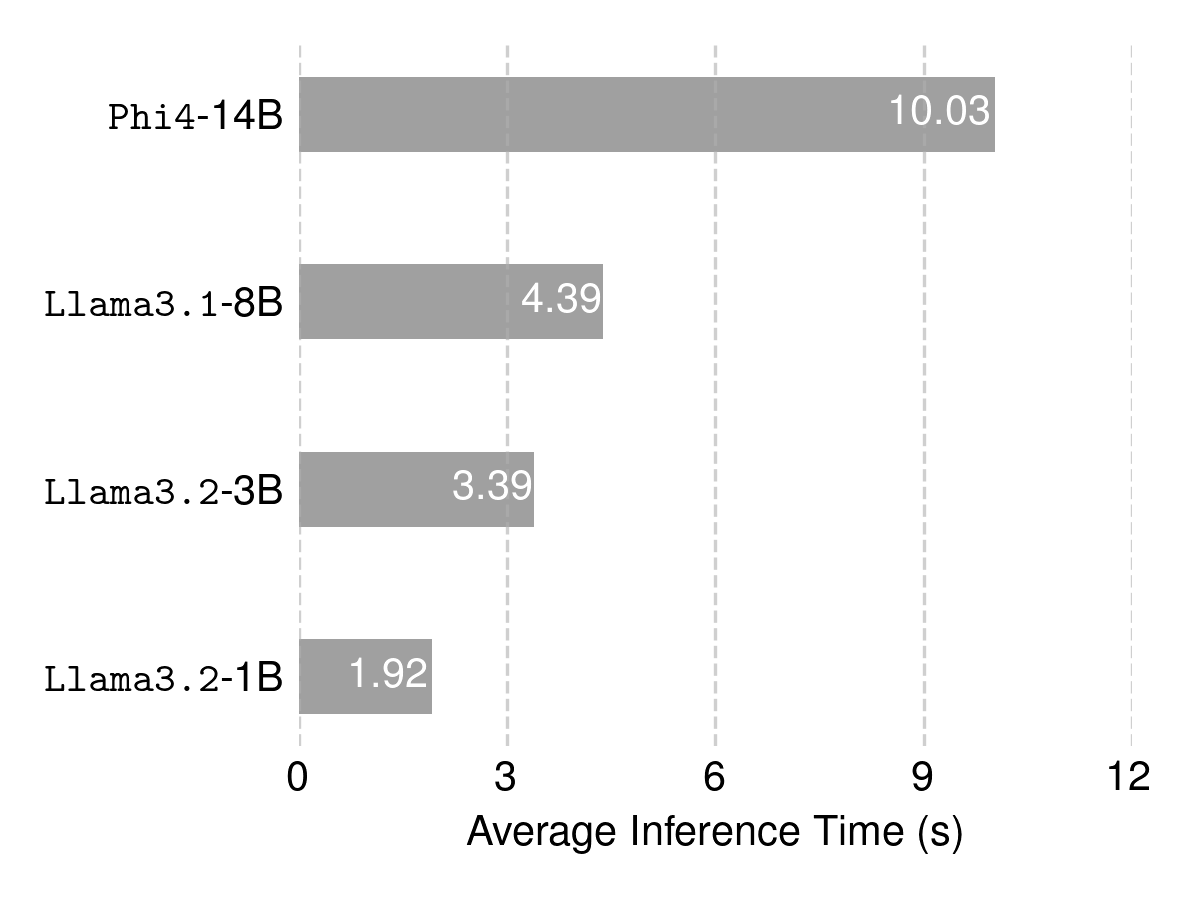}
\caption{Average LLM inference time (A100 GPU).}
\label{fig:inf_time_task2}
\vspace{-.5em}
\end{figure}

For Task (1) BSSID Selection, the STA must make roaming decisions within $10$--$100\,\mathrm{ms}$ after a scan to meet near-real-time (near-RT) requirements. However, as shown in Fig.~\ref{fig:inf_time_task2}, current LLMs exhibit inference times on the order of seconds, far exceeding the required latency. Furthermore, frequent inferences add to computational overhead and power consumption.

Thus, although LLMs provide powerful in-context learning, their current latency makes them more suitable for tasks that require less frequent, non-real-time inference in practical on-device deployments.

\section{Task (2): Online Roaming Optimization by On-Device LLM} \label{sec:task2}
Having established that an LLM can effectively choose APs, we now adapt it to the challenge of deciding \emph{when} to roam in on-device setup. Our second objective is to evaluate whether an on-device LLM can enhance Wi-Fi roaming by dynamically adjusting the roaming threshold in on-device setup. This task requires less frequent and time-critical inference, making it well-suited for practical on-device deployment.

\subsection{Problem Statement: Dynamic Threshold Selection}
A major challenge in Wi-Fi roaming is determining \emph{when} to initiate handover. This is typically controlled by a fixed RSSI threshold (\texttt{scanRSSI}), below which STA begins scanning for alternative APs. However, as previously discussed, a single static value can lead to two undesirable outcomes: setting it too high causes overly frequent scans and roaming (wasting energy and risking ping-pong effects), whereas setting it too low risks lingering on a deteriorating link until performance degrades or the connection drops. Building on the context-aware approach in Section~\ref{sec:task1}, we now aim to dynamically select the threshold in real time, leveraging 
an \emph{on-device LLM}.

Yet deploying a large LLM directly on mobile or edge platforms is challenging due to hardware constraints such as limited memory, battery capacity, and compute power. Accordingly, one needs to find a way to reduce the LLM’s computation overhead while preserving strong performance in threshold‐selection decisions.

\subsection{Approach: Adaptive Threshold via On-Device LLM}

\BfPara{Cross-Layer Operation}
On-Device LLM operates within the application layer but directly controls MAC-layer parameters, specifically the roaming threshold. This illustrates a concrete example of \emph{application-layer AI for lower-layer wireless control}, where high-level reasoning guides low-level network decisions.

\BfPara{Model Selection and Quantization}
Deploying large models on resource-constrained edge devices requires careful consideration of model size and computational resources. To ensure practical lower-layer operation, we select a model size that optimally balances inference speed with decision accuracy. 
We further reduce computational overhead by employing the \texttt{Q2\_K} quantization scheme. Specifically, \texttt{Q2\_K} compresses weights into groups of $16$, encoding each weight with $2$-bit precision and using one shared $4$-bit scale and offset per block. This compact encoding needs only \(2.56\) bits per parameter—about \(8\times\) fewer bits than \texttt{FP16}—while staying fully compatible with the \texttt{GGUF} model format supported by \texttt{llama.cpp}.

\BfPara{Task-Oriented Optimization}
Since Task~(2) is focused specifically on determining \emph{when} to initiate roaming, rather than choosing a specific AP, our optimization objective is explicitly designed around dynamic threshold adjustment. We carefully tune the frequency at which the LLM updates roaming thresholds, enabling it to react quickly to significant RSSI fluctuations while minimizing computational and energy overhead.

\subsection{Experiments}
Unless otherwise noted, we evaluate this approach under the same general setup, baselines, and metrics used for Task~(1) in Sec.~\ref{sec:evaluation}. Detailed experimental setups and parameter settings are provided in Table~\ref{tab:setup} in Appendix~\ref{sec:setup}.

\BfPara{Effect of Quantization}
Q4\_K\_MTable~\ref{table:quantization_task2} compares the impact of various quantization schemes (\eg \texttt{Q2\_K}, \texttt{Q3\_K\_M}, \texttt{Q4\_K\_M}) on model size and roaming performance. Quantization significantly reduces the model's footprint from 8.5,GB (\texttt{Q8\_0}) down to as low as 3.2,GB (\texttt{Q2\_K}), while maintaining comparable performance metrics (approximately $\approx138–145$ handovers and around $\approx-58$,dBm RSSI). Given these results, we select \texttt{Q2\_K} quantization for subsequent experiments, as it provides substantial resource savings with negligible loss in decision accuracy.

\begin{table}[!h]   
\centering
\resizebox{.95\columnwidth}{!}{
\centering
\begin{tabular}{c|c c|c}
\toprule[1pt]
\textbf{Quant. Method} & \textbf{\textit{\# HO}} & \textbf{\textit{AvgRSSI}}~(dBm) & \textbf{Model Size}~(GB)\\
\midrule
\xmark    & 143 & -57.98 & 16 \\
\cmidrule(lr){1-4}
\texttt{Q8\_0}    & 140 & -58.62 & 8.5\\
\texttt{Q6\_K}    & 139 & -58.63 & 6.6\\
\texttt{Q5\_K\_M} & 144 & -58.26 & 5.7\\
\texttt{Q4\_K\_M} & 145 & -58.80 & 4.9\\
\texttt{Q3\_K\_M} & 138 & -58.89 & 4.0\\
\textbf{\texttt{Q2\_K}}    & 138 & -58.85 & 3.2\\
\bottomrule[1pt]
\end{tabular}


}
\caption{Comparison of quantization schemes for the \texttt{Llama3.1-8B} model in terms of roaming decision and memory footprint \cite{grattafiori2024llama}.}
\label{table:quantization_task2}
\vspace{-.5em}
\end{table} 
The adoption of \texttt{Q2\_K} reduces the model size by approximately 2.7×, from 8.5 GB with \texttt{Q8\_0} quantization to 3.2 GB, enabling efficient operation within the 16,GB unified memory capacity of the Apple M-series MacBook used for our on-device experiments.

\BfPara{LLM Model Comparison}
Table~\ref{table:modelselection_task2} further compares models ranging from 1B to 14B parameters. Larger variants (\eg \texttt{Phi4}-14B \cite{abdin2024phi4technicalreport}) provide slightly higher RSSI (around \(-57.9\)\,dBm) but yield more handovers (156) and longer inference times (up to 46.9\,s). Smaller models (\eg 1B) drastically reduce inference delays (4.4\,s) and handovers (124) but at the cost of a weaker average signal (\(-60.07\)\,dBm). 
Models in the 3–8B range often offer a balanced trade-off between speed and roaming performance.

\begin{table}[!h]   
\centering
\resizebox{.95\columnwidth}{!}{
\centering
\begin{tabular}{c|c c|c}
\toprule[1pt]
\textbf{Quant. Method} & \textbf{\textit{\# HO}} & \textbf{\textit{AvgRSSI}}~(dBm) & \textbf{Model Size}~(GB)\\
\midrule
\texttt{Llama3.2}-1B & 124 & -60.07 & 0.6 \\
\texttt{Llama3.2}-3B & 150 & -58.47 & 1.4 \\
\texttt{Llama3.1}-8B & 138 & -58.85 & 3.2 \\
\texttt{Phi4}-14B   & 156 & -57.90 & 5.5 \\
\bottomrule[1pt]
\end{tabular}


}
\caption{Comparing LLMs for dynamic threshold selection (Task~2).}
\label{table:modelselection_task2}
\vspace{-.5em}
\end{table} 

\BfPara{How Often Should the LLM Adjust the Threshold?}
We next vary the frequency of roaming threshold adjustments (from 10\,s to 300\,s). Decreasing this interval to 10\,s increases the number of roaming (about 147) while boosting average RSSI (to \(-57.86\)\,dBm). Increasing the interval to 300\,s cuts roamings (down to 119) but weakens the average RSSI (to \(-59.94\)\,dBm). 
A 30\,s interval emerges as a practical middle ground, 
balancing roaming stability, signal quality, and computational overhead.

\begin{figure}[!h]
\centering
\includegraphics[width=1.\linewidth]{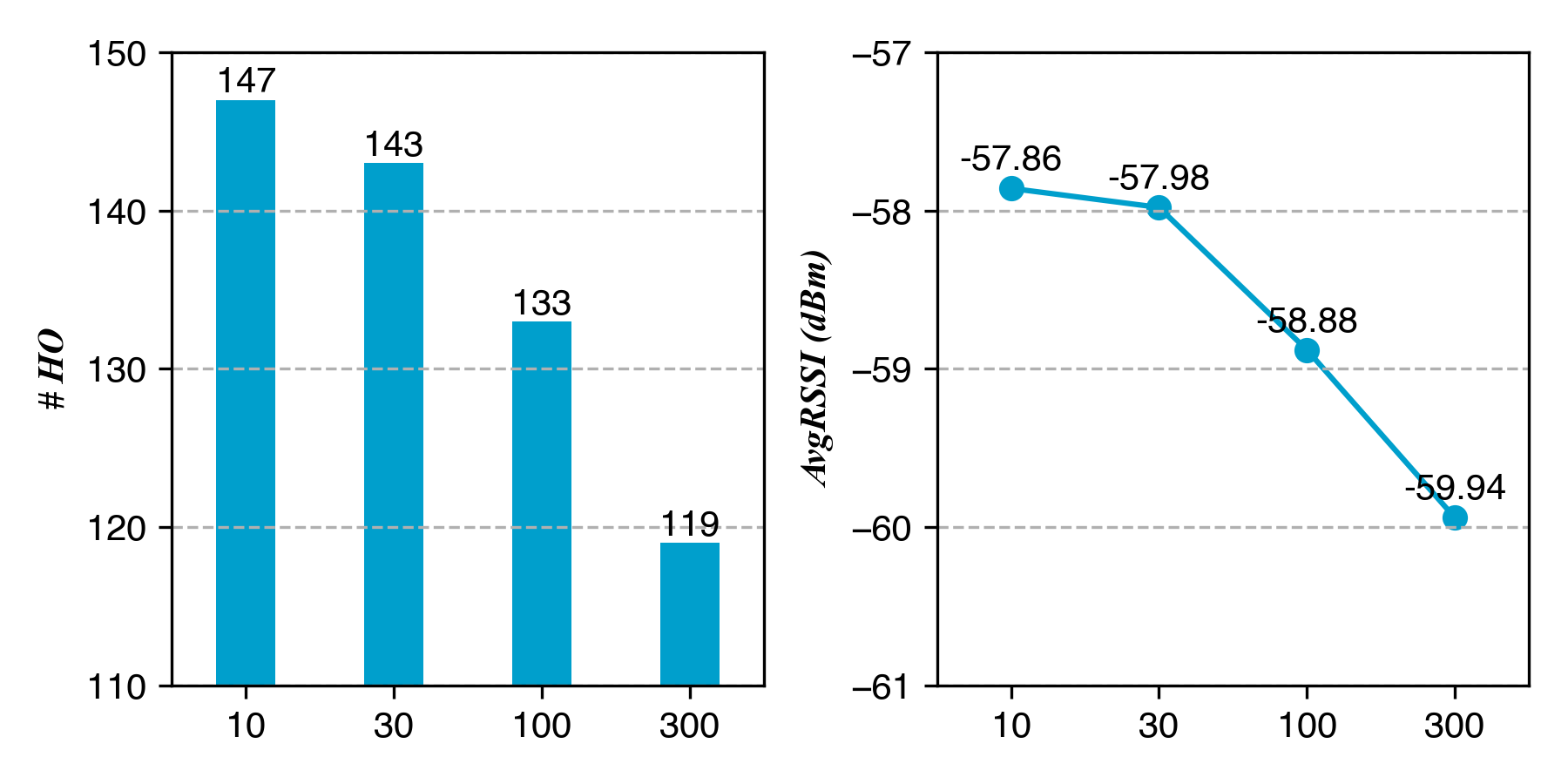}
\caption{Frequency of threshold adjustments and its effect on performance.}
\vspace{-.5em}
\end{figure}

\begin{figure*}[!h]
\vspace{-.5em}
\centering
\includegraphics[width=.5\linewidth]{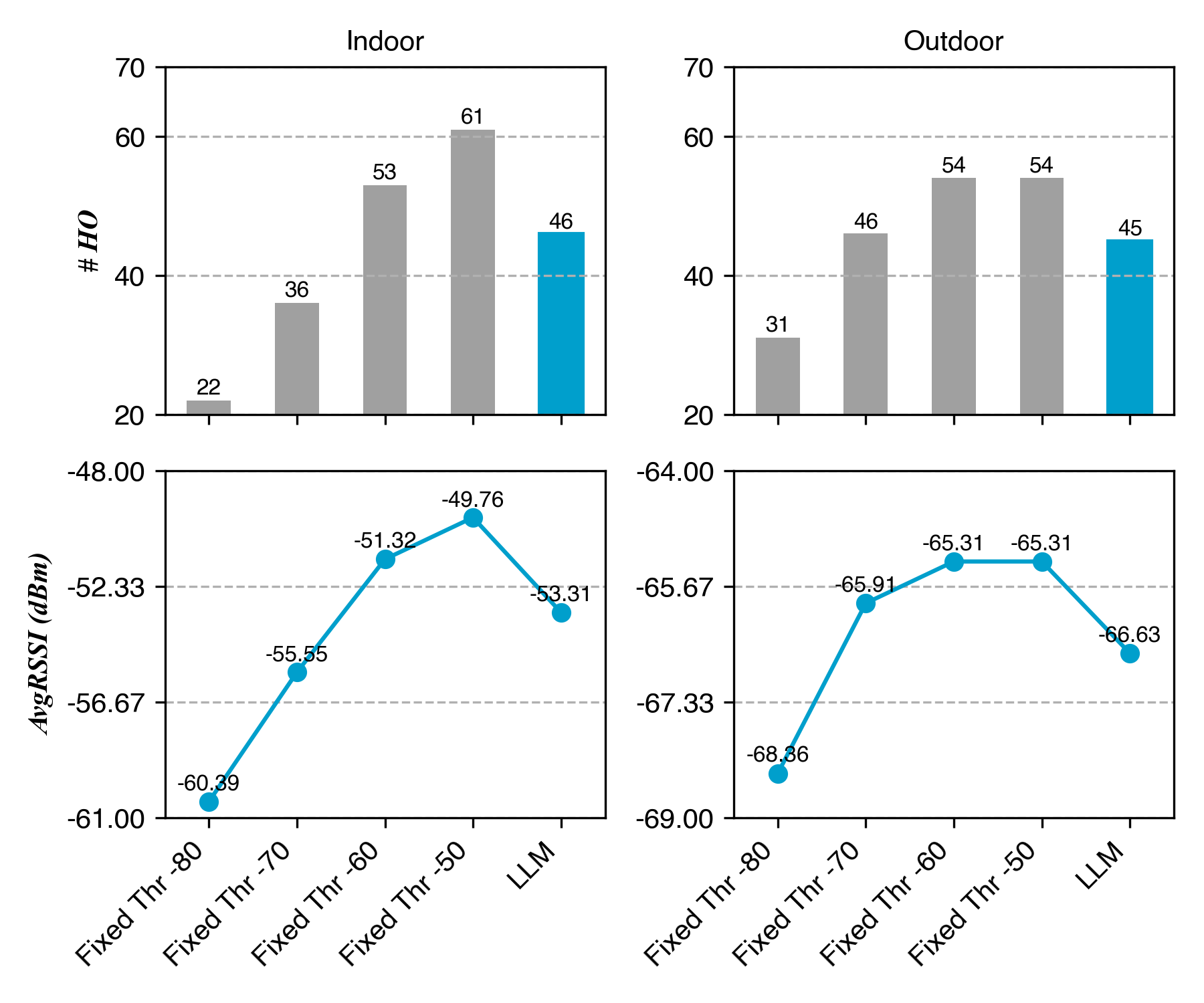}
\vspace{-.5em}
\caption{Legacy vs.\ Context-Aware On-Device LLM: Comparison of \textbf{\textit{\# HO}} (top) and \textbf{\textit{AvgRSSI}} (bottom) for each method for dynamic threshold selection (Task 2) in various roaming scenarios.}
\label{fig:comparison_task2}
\vspace{-1.em}
\end{figure*}

\BfPara{Comparison on \emph{On-Device}: Legacy vs. LLM}
Finally, we compare our on-device LLM against four fixed thresholds 
(\(-50, -60, -70, -80\)\,dBm) in on-device setup. As illustrated in Fig.~\ref{fig:comparison_task2}, each static threshold performs best under specific conditions yet fails to generalize. 
For instance, \(-50\)\,dBm consistently achieves a strong average RSSI but induces excessive handovers; \(-80\)\,dBm cuts handover frequency but suffers from weaker connectivity. Intermediate thresholds like \(-60\)\ or \(-70\)\,dBm offer varied performance trade-offs, often performing well in outdoor environments but less consistently indoors. 

In contrast, the proposed context-aware on-device LLM dynamically adjust its threshold to local context information, enabling it to maintain balanced RSSI performance and moderate handover frequency across diverse scenarios. While it does not always achieve the absolute best performance on every single metric, the LLM notably delivers robust, generalized performance without manual tuning. This underscores the practical value and effectiveness of cross-layer, adaptive threshold optimization using on-device LLMs for real-world Wi-Fi roaming scenarios.


\section{Discussion} \label{sec:discussion}
In conclusion, this section discusses key insights and practical considerations for deploying LLMs as edge-level decision-making agents for wireless lower-layer operations through the following key questions:

\BfPara{Q1: \textit{Does context information improve Wi-Fi roaming decisions?}}  
Incorporating real-time context—such as location and time—enables LLMs to make more informed roaming decisions. However, as indicated in Table~\ref{tab:ablation_contextinfo}, using every available context feature does not necessarily yield optimal results. Instead, selectively incorporating task-relevant context can lead to better performance, highlighting the importance of carefully choosing which contextual inputs are most beneficial.

\begin{table}[!h]   
\centering
\resizebox{1.\columnwidth}{!}{
\centering
\newcolumntype{R}{>{\raggedleft\arraybackslash}X}
\begin{tabular}{l|ccc|cc}
\toprule[1.5pt]
\textbf{Combination} & \textbf{Location} & \textbf{Time} & \textbf{Battery} & \textbf{\textit{\# HO}} & \textbf{\textit{AvgRSSI}}~(dBm) \\
\midrule[.8pt]
w/o Context & \xmark & \xmark & \xmark & 151 & -58.03 \\
Time + Battery & \xmark & \cmark & \cmark & 142 & -58.19 \\
Location + Battery & \cmark & \xmark & \cmark & 142 & -58.19 \\
\textbf{Location + Time} & \cmark & \cmark &  \xmark & 143 & -57.98 \\
{w/ All} & \cmark & \cmark & \cmark & 146 & -58.05 \\
\bottomrule[1.5pt]
\end{tabular}

}
\caption{Ablation study for context information.}
\label{tab:ablation_contextinfo}
\end{table} 

\BfPara{Q2: \textit{Why use an On-Device LLM instead of conventional ML models?}}  
On-device LLMs excel at zero-shot and few-shot learning, enabling generalization to new environments without extensive retraining, thanks to their inherent in-context learning capabilities. They can dynamically integrate multiple contextual inputs, adapting seamlessly to diverse scenarios. By contrast, conventional ML models—such as those based on DRL or supervised learning—typically rely on fixed heuristics and require substantial, environment-specific retraining (see Table~\ref{tab:llm_vs_ml}).

\BfPara{Q3: \textit{Is an on-device LLM practical for lower-layer operations?}}
As discussed in Sec.~\ref{sec:limitation}, current on-device LLM implementations exhibit inference latencies on the order of seconds, posing challenges for strict near-real-time (RT) operations. Nevertheless, recent advances in dedicated AI accelerator hardware are expected to significantly reduce inference times, greatly enhancing the practicality of on-device LLMs even for near-RT Wi-Fi tasks such as roaming. Moreover, for scenarios like threshold adjustment—where decisions are required less frequently and slightly higher latencies can be tolerated—the existing on-device LLM deployments already demonstrate substantial feasibility. Importantly, although inference occurs at the application layer, the decisions made directly adjust MAC-layer parameters, clearly illustrating the viability of \emph{application-layer AI enabling adaptive lower-layer wireless control}. Demonstrations and details are available at \href{https://github.com/abman23/on-device-llm-wifi-roaming}{github.com/abman23/on-device-llm-wifi-roaming}.

\BfPara{Q4: \textit{What are the main challenges and future directions?}}
Deploying on-device LLMs presents several critical challenges, particularly managing computational overhead and optimizing models for strict hardware constraints inherent to edge devices. Although current on-device LLMs show promising capabilities, their universal practicality for all lower-layer wireless operations—especially those demanding strict real-time responsiveness—remains limited. Future research must focus on refining open-source LLM architectures and techniques, emphasizing efficiency improvements in latency, memory usage, and power consumption. Additionally, investigating the broader applicability of on-device LLMs to other essential wireless control tasks beyond Wi-Fi roaming will be crucial in fully harnessing their potential as versatile and practical AI-based decision makers in wireless communication systems.

\newpage
\bibliography{refs}
\bibliographystyle{icml2025}

\newpage
\appendix
\onecolumn
\appendix
\section{Real-World Demonstration} \label{sec:demo}
\begin{figure*}[!ht]
    \centering  
    \begin{subfigure}[t]{.47\linewidth}
        \centering  
        \includegraphics[width=\linewidth]{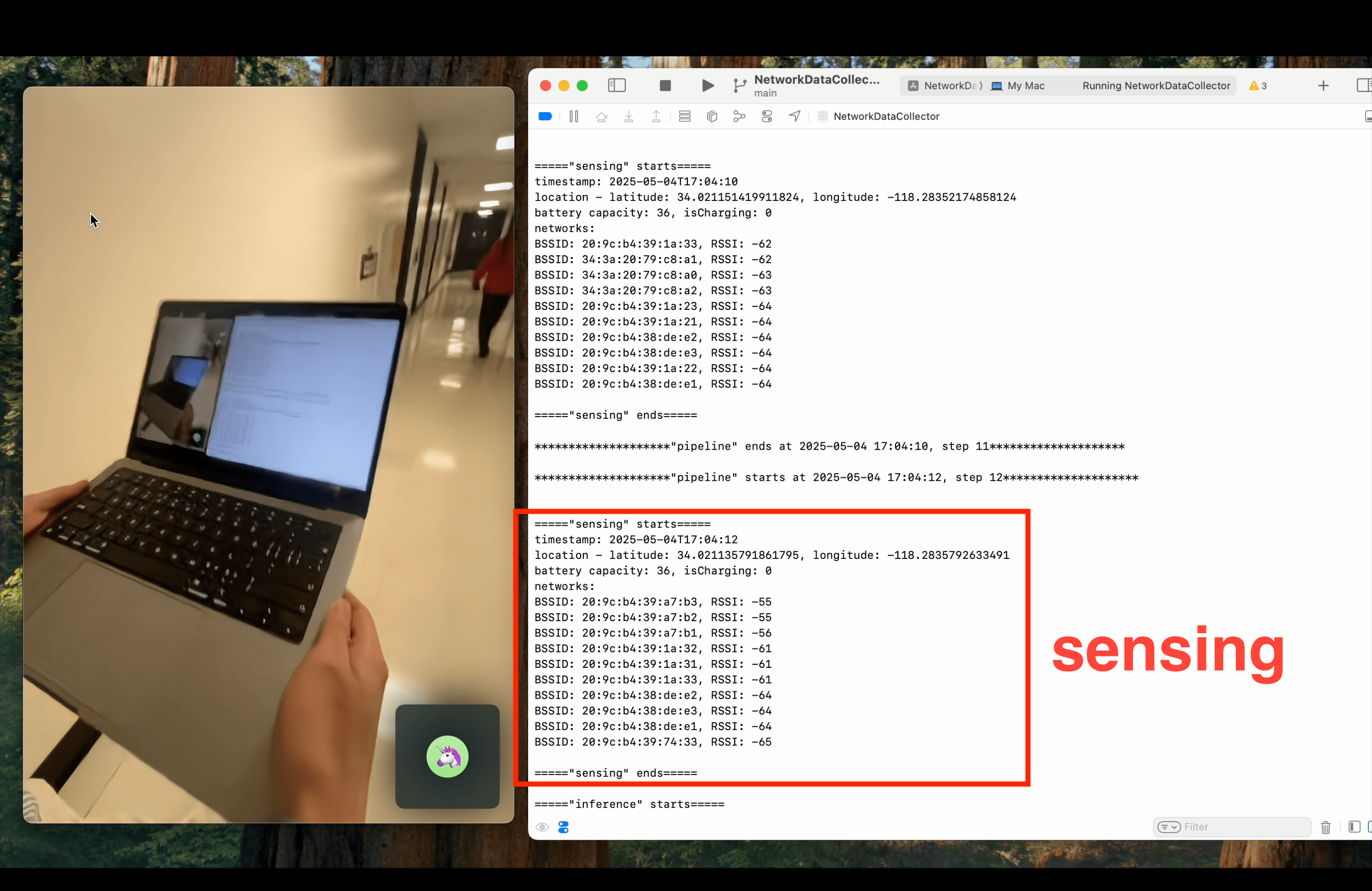}
        \caption{Indoor — sensing phase}
        \label{fig:indoor-task2-sensing}
    \end{subfigure}\hspace*{.015\textwidth}%
    \begin{subfigure}[t]{.47\linewidth}
        \centering
        \includegraphics[width=\linewidth]{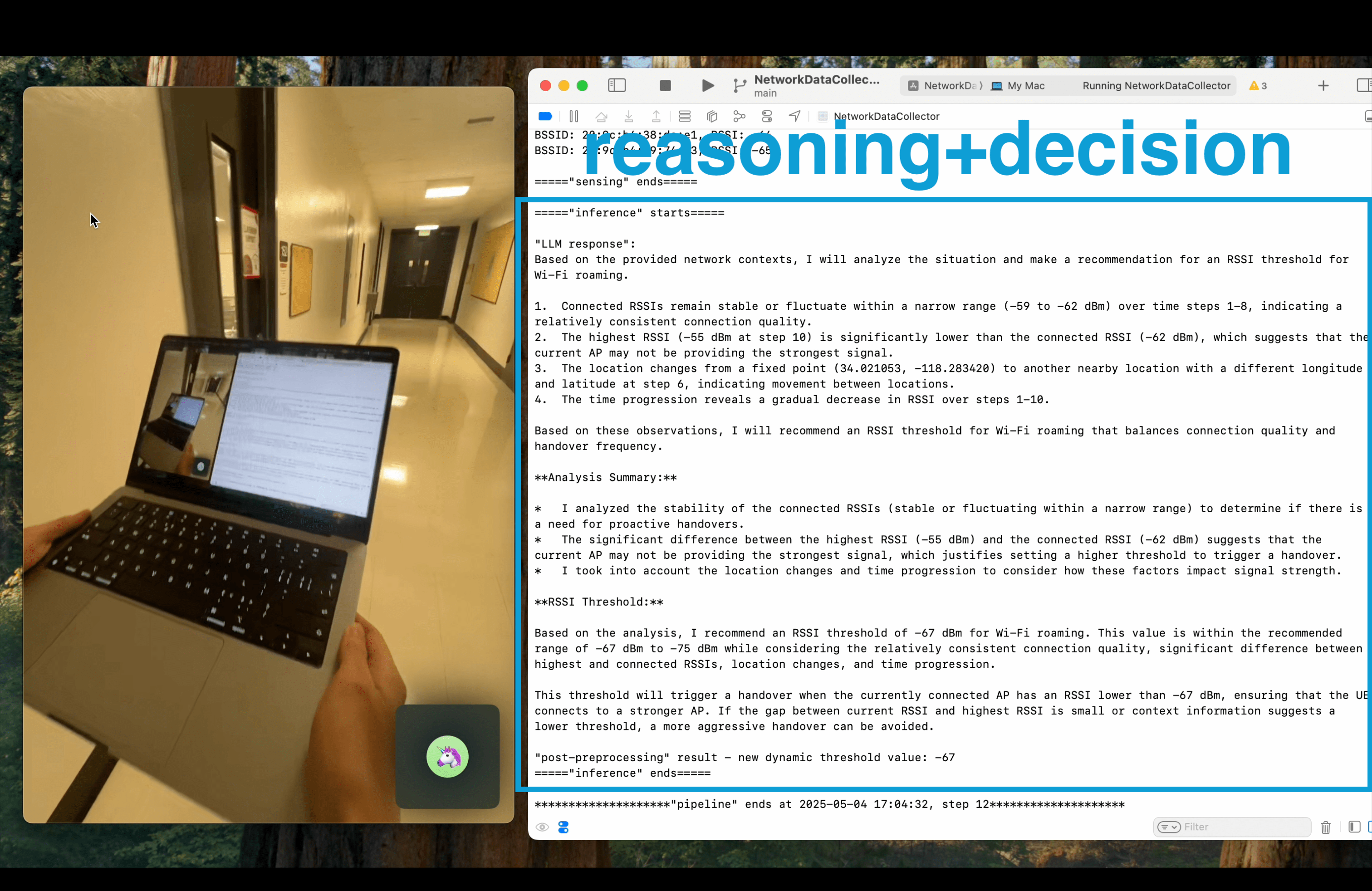}
        \caption{Indoor — reasoning \& decision}
        \label{fig:indoor-task2-reasoning+decision}
    \end{subfigure}\hspace*{.015\textwidth}%
    \\ \vspace*{.1\textwidth}%
    \begin{subfigure}[t]{.47\linewidth}
        \centering  
        \includegraphics[width=\linewidth]{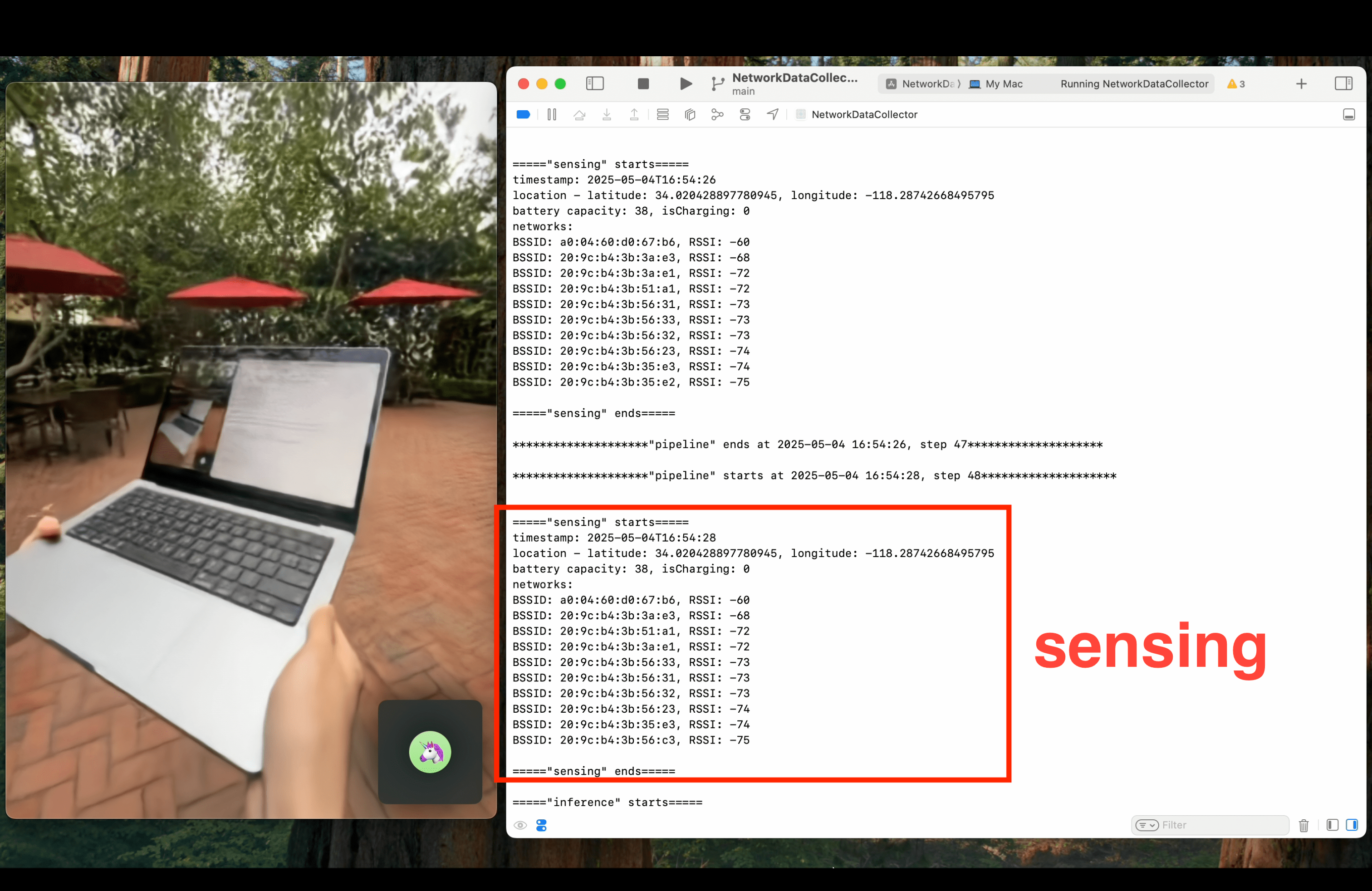}
        \caption{Outdoor — sensing phase}
        \label{fig:outdoor-task2-sensing}
    \end{subfigure}\hspace*{.015\textwidth}%
    \begin{subfigure}[t]{.47\linewidth}
        \centering
        \includegraphics[width=\linewidth]{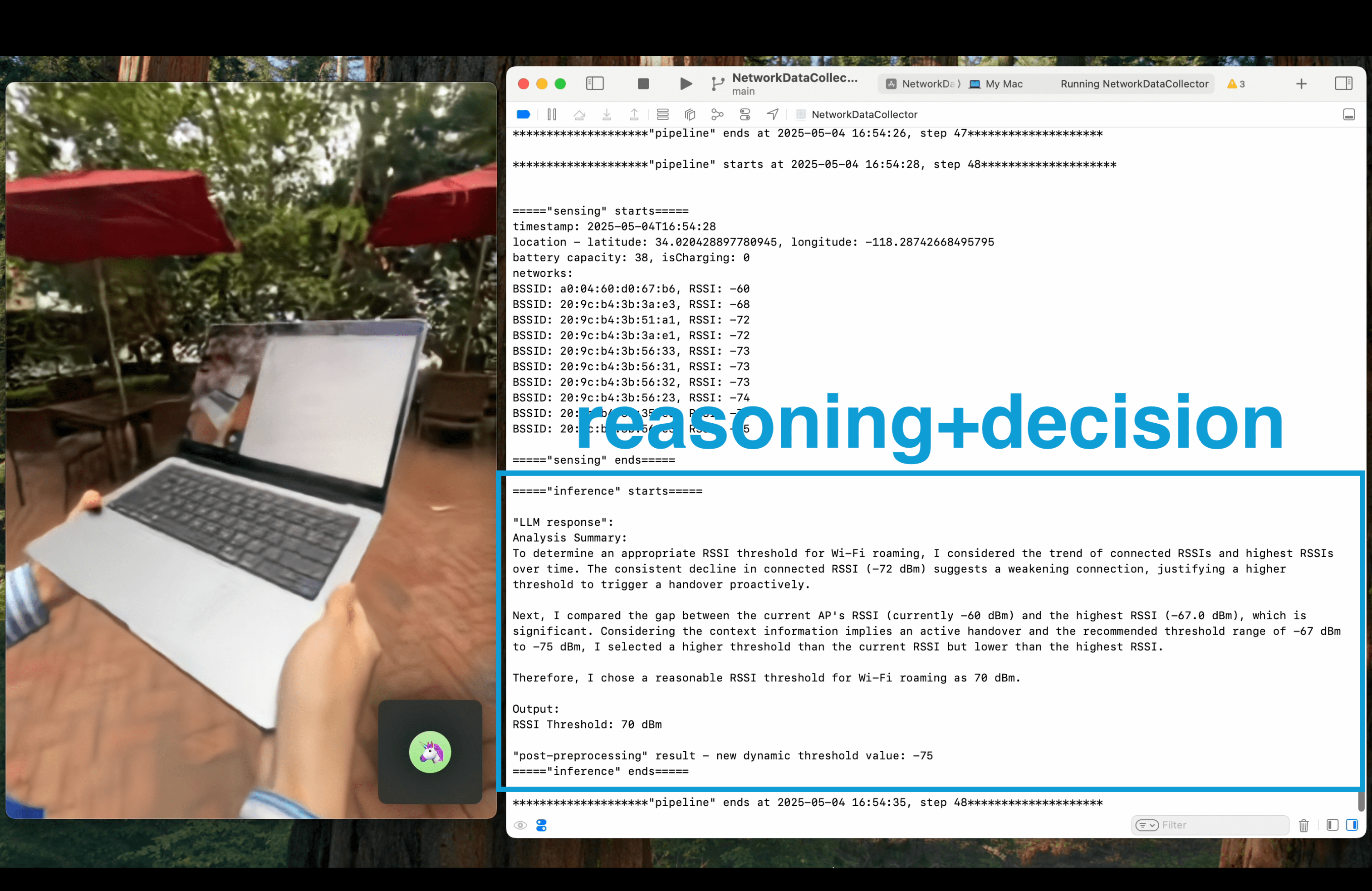}
        \caption{Outdoor — reasoning \& decision}
        \label{fig:outdoor-task2-reasoning+decision}
    \end{subfigure}\hspace*{.015\textwidth}%
\caption{Real-world demonstration of our on-device LLM for \textbf{Task 2 (dynamic threshold adjustment)}.  
  In each setting the laptop (Macbook Pro) first \emph{senses} RSSI and context information (left), then the LLM \emph{reasons} and outputs an adaptive roaming threshold (right), validating practical operation indoors and outdoors.}
\label{fig:demo}
\end{figure*}

To validate on-device deployment, we ran our LLM-based roaming agent on a MacBook Pro (Apple M-series, 16 GB RAM, \texttt{macOS}).  
Figure~\ref{fig:demo} captures key moments from a live screen-recorded session: the sensing phase (RSSI and context collection) followed by the LLM’s reasoning and threshold update, shown for both indoor and outdoor walks. Comprehensive performance metrics confirm the feasibility of real-time operation on consumer-grade hardware.  
A full demonstration video is available at \href{https://github.com/abman23/on-device-llm-wifi-roaming}{github.com/abman23/on-device-llm-wifi-roaming}.

\newpage
\section{Qualitative Assessment of On-Device LLM Reasoning}\label{sec:reasoning}
\begin{figure*}[!ht]
    \centering  
    \includegraphics[width=.95\linewidth]{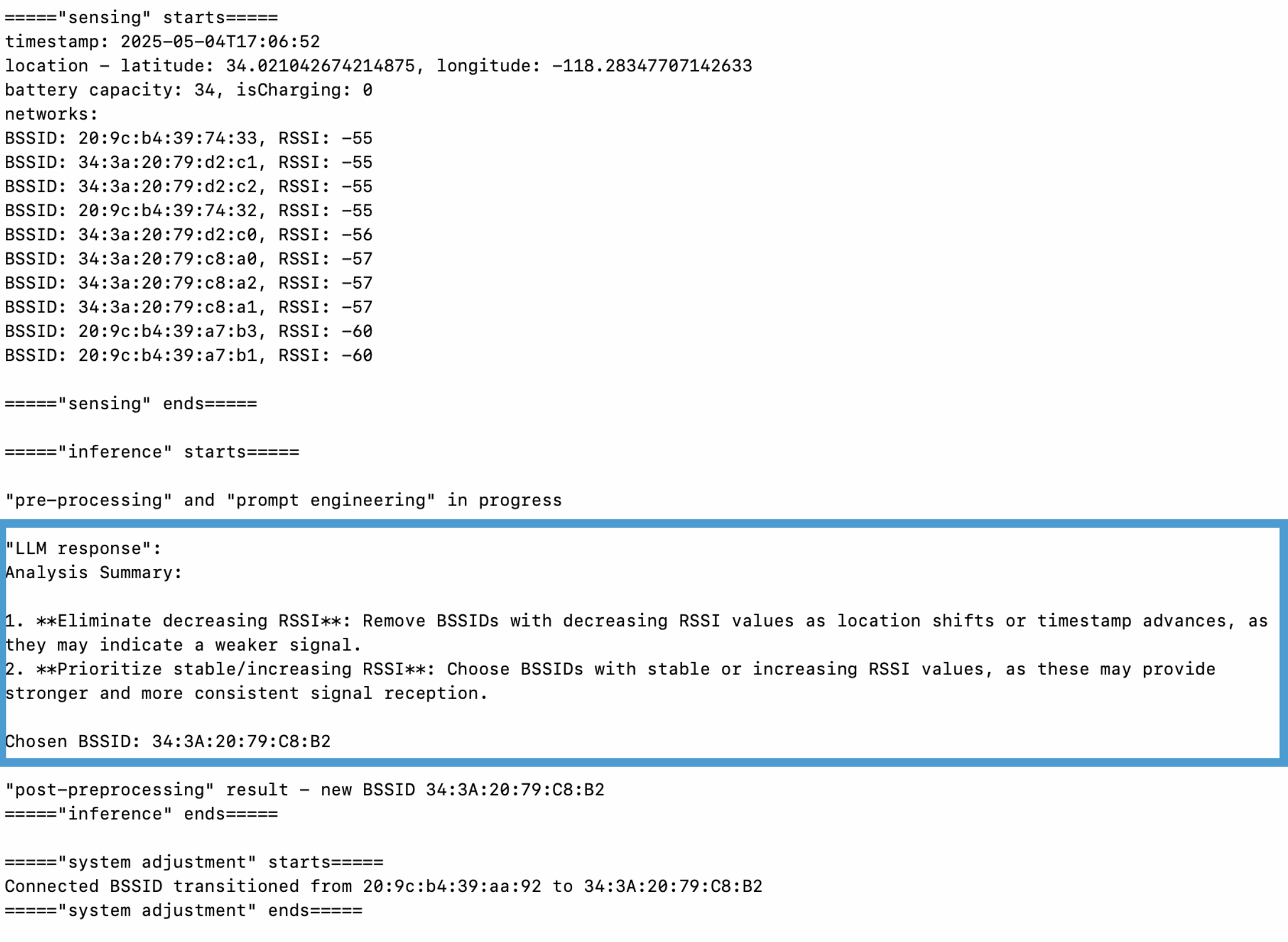}
    \caption{Reasoning output of the on-device LLM during an indoor roaming scenario. The model assesses AP candidates by eliminating those with declining RSSI values and prioritizes APs with stable signal strength, ultimately selecting BSSID \texttt{34:3A:20:79:C8:B2} for handover.}
    \label{fig:reasoning}
    \vspace{.5em}
\end{figure*}

In the real-world indoor demonstration shown in Fig.~\ref{fig:reasoning}, the on-device LLM demonstrates reasoning aligned with practical wireless network management strategies. The inference process reflects logical heuristics such as filtering out APs exhibiting decreasing signal strength and prioritizing stable or improving RSSI, consistent with best practices for effective roaming decisions. This reasoning is derived purely from local contextual (\eg timestamp, location) and Wi-Fi measurement (\eg RSSI-BSSID pairs) information. The correctness of the LLM’s reasoning is validated through the subsequent successful handover, evidenced by system logs. This qualitative evaluation complements the quantitative results presented earlier (\S\ref{sec:task2}), reinforcing the capability of on-device LLMs to execute valid, interpretable reasoning suitable for lower-layer wireless control at edge-device.

\newpage
\section{Setup}\label{sec:setup}
Table~\ref{tab:setup} details the hardware, dataset statistics, and model-tuning parameters used in \textbf{Task 1} (AP selection) and \textbf{Task 2} (threshold adjustment).  
All real-world experiments ran on a consumer-grade MacBook Pro with an Apple M-series processor (16 GB RAM, \texttt{macOS}), demonstrating that our on-device LLM pipeline is feasible on typical edge hardware.

\begin{table}[!h]   
\centering
\resizebox{.6\columnwidth}{!}{
\begin{tabular}{ll}
\toprule[1pt]
\textbf{HW / SW} & \\
\midrule
Device                   & MacBook Pro (M3, 16GB) \\
OS                       & \texttt{macOS} \\
Wi-Fi Specification      & \texttt{IEEE 802.11ax} (Wi-Fi 6E) \\
Wi-Fi Library            & \texttt{CoreWLAN} \\
\midrule
\textbf{Dataset (Wi-Fi Log)} & \\
\midrule
Scenarios                & Indoor, Outdoor \\
Time Stamps per Scenario & 1944, 5808, 1944 \\
Sampling Interval        & 1 sec \\
Training : Test Split    & 80\% : 20\% \\
\midrule
\textbf{Task Configuration} & \\
\midrule
\texttt{scanRSSI} Threshold & -70 dBm \\
Contextual Inputs        & Location + Time \\
Logs per Handover Decision & 10 \\
Threshold Adjustment Interval (Task 2) & 30 sec \\
\midrule
\textbf{On-Device LLM Configuration} & \\
\midrule
Base Model               & \texttt{Llama-3.1-8B-Instruct} \\
Quantization Scheme      & \texttt{Q2\_K} \\
\midrule
\textbf{Fine-Tuning Configuration} & \\
\midrule
Learning Rate            & $2\times10^{-4}$ \\
Batch Size / Grad. Accumulation & 4 / 4 \\
Number of Epochs         & 1 \\
Weight Decay             & 0.01 \\
LoRA Rank / Alpha        & 128 / 128 \\
Optimizer                & \texttt{AdamW} \\
\bottomrule[1pt]
\end{tabular}






}
\caption{Experimental setup and parameter settings for Tasks 1 \& 2.}
\label{tab:setup}
\end{table} 

\newpage
\section{Data Collection} \label{sec:data}
\begin{figure*}[!h]
  \centering
  \begin{subfigure}[t]{.25\linewidth}
    \centering\includegraphics[width=\linewidth]{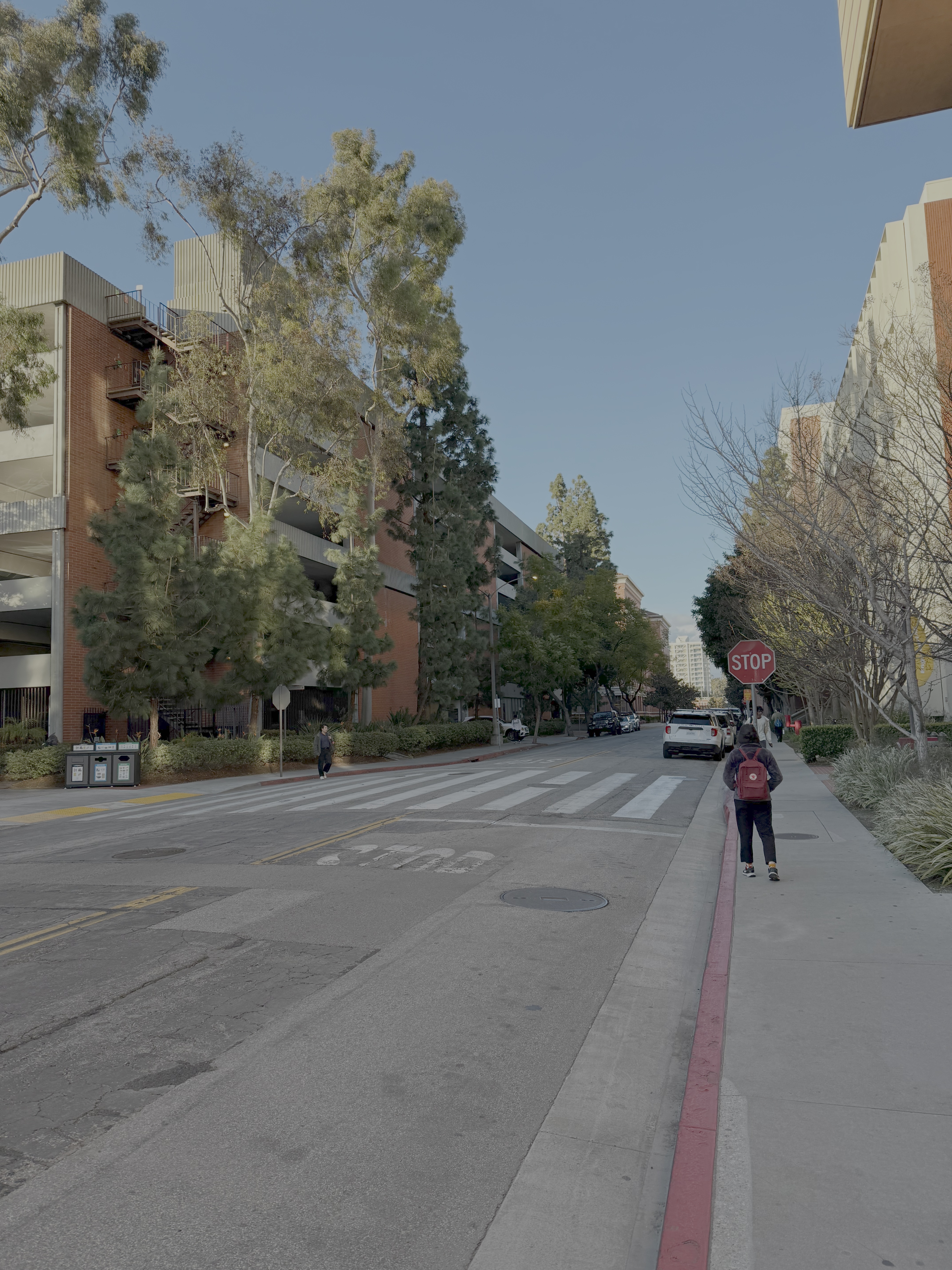}
    \caption{Outdoor scene (campus walkway)}
    \label{fig:outdoor}
  \end{subfigure}\hspace*{.025\textwidth}%
  \begin{subfigure}[t]{.25\linewidth}
    \centering\includegraphics[width=\linewidth]{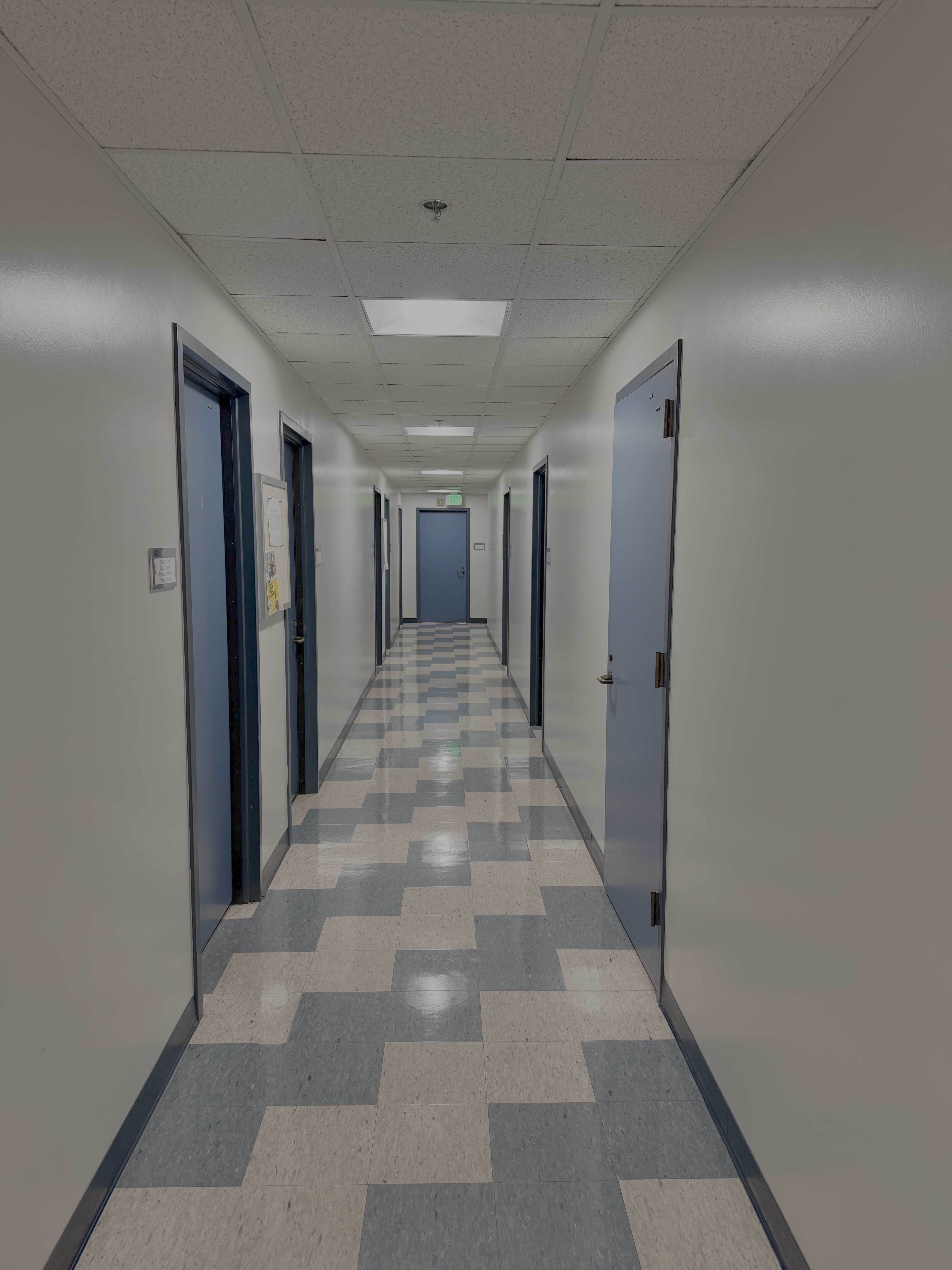}
    \caption{Indoor scene (corridor)}
    \label{fig:indoor}
  \end{subfigure}\hspace*{.025\textwidth}%
  \begin{subfigure}[t]{.25\linewidth}
    \centering\includegraphics[width=\linewidth]{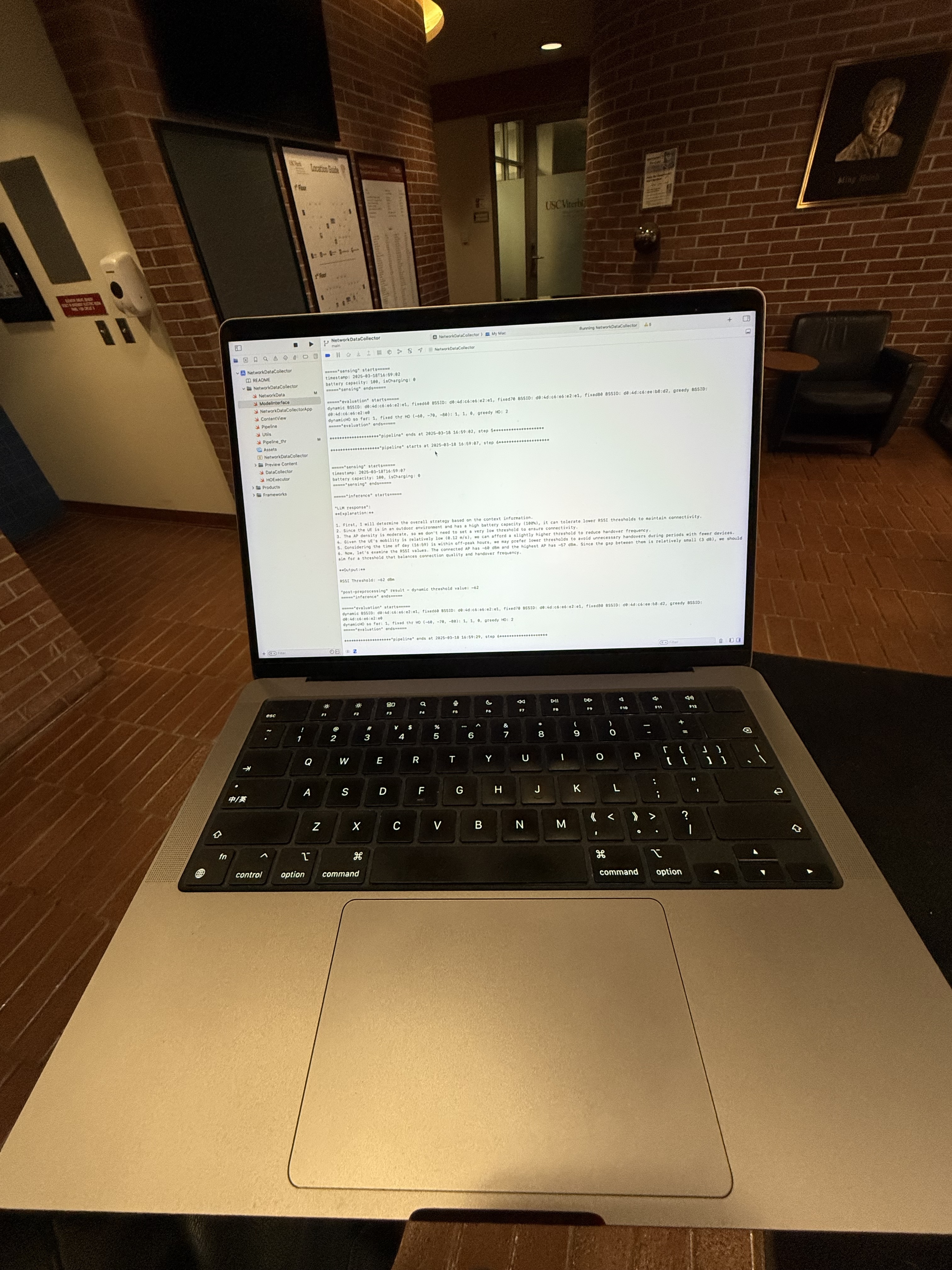}
    \caption{Data-capture rig (MacBook Pro)}
    \label{fig:datacollection_setup}
  \end{subfigure}
  \caption{Data-collection environments used for on-device LLM evaluation.}
  \label{fig:datacollection}
\end{figure*}

We collected Wi-Fi roaming datasets from diverse indoor and outdoor environments, as illustrated in Fig.~\ref{fig:datacollection}. Each of the ten sessions recorded approximately \(\sim\!1{,}000\) consecutive data points, captured at one-second intervals. Each data point captures detailed information, including:

\begin{itemize}[leftmargin=*]
  \item \textbf{AP scan:} BSSID, RSSI, etc.
  \item \textbf{Device context:} timestamp, geographical coordinates (latitude and longitude), battery status, etc.
\end{itemize}
The resulting dataset underpins both post-training and evaluation of our on-device LLM.

\end{document}